\documentclass[conference,a4paper]{IEEEtran}
\IEEEoverridecommandlockouts
\usepackage{cite}
 \usepackage{color}

\def\blue{\textcolor{blue}}
\usepackage{float}
\usepackage{derivative}
\usepackage{amsmath,amssymb,amsfonts}
\usepackage[english]{babel}
\usepackage{amsthm}
\theoremstyle{definition}

\usepackage{graphicx}
\usepackage{caption}
\usepackage{subcaption}
\usepackage{multirow}
\usepackage[left=1.35cm,right=1.35cm,top=1.9cm,bottom=4.1cm]{geometry}
\usepackage{textcomp}
\usepackage{xcolor}
\usepackage{bbm}
\usepackage{algorithm}
\usepackage{algpseudocode}

\def\BibTeX{{\rm B\kern-.05em{\sc i\kern-.025em b}\kern-.08em
    T\kern-.1667em\lower.7ex\hbox{E}\kern-.125emX}}
\usepackage{tikz}

\begin{document}
\def\blue{\textcolor{blue}}
\title{Graph Neural Networks for Resource Allocation in Integrated Sensing and Communication \\

}
\newcommand\blfootnote[1]{%
  \begingroup
  \renewcommand\thefootnote{}\footnote{#1}%
  \addtocounter{footnote}{-1}%
  \endgroup
}
\author{\IEEEauthorblockN{Junyu Liao$^{\ast}$, Lili Chen$^{\ast}$$^{\dagger}$, Tian Han$^{\ast}$, and Jamie Evans$^{\ast}$}

\IEEEauthorblockA{$^{\ast}$Department of Electrical and Electronic Engineering, University of Melbourne, Australia \\
$^{\dagger}$Department of Electrical and Computer Systems Engineering, Monash University, Australia\\
Email: junyu.liao@student.unimelb.edu.au, \{tian.han1, jse\}@unimelb.edu.au, lili.chen@monash.edu\\
}

}
\maketitle
\begin{abstract}

In vehicular integrated sensing and communication (ISAC) networks, each service-provider vehicle uses a limited power budget to serve multiple sensing targets and communication receivers. Maximizing the communication sum-rate subject to sensing signal-to-interference-plus-noise ratio (SINR) constraints is nonconvex, and iterative optimization is costly to rerun for every network layout. This paper proposes a graph neural network (GNN)-based power allocation algorithm, termed GNN-ISAC. Concretely, GNN-ISAC addresses the aforementioned problem by considering two distinct edge types in the graph representation, one capturing the inter-provider interference, the other representing the intra-provider power-budget sharing. GNN-ISAC is trained without labels using a sensing-penalized sum-rate loss. We also adapt the weighted mean-square error minimization (WMMSE) algorithm to this problem by introducing a sensing-power reservation step. Simulation results show that GNN-ISAC achieves communication sum-rates close to WMMSE with low sensing-link violation rates and much less computation. It also outperforms an existing heterogeneous GNN and a channel-blind baseline, and generalizes to larger networks without retraining.
\end{abstract}

\begin{IEEEkeywords}
Graph neural networks, integrated sensing and communication, power allocation, vehicular networks
\end{IEEEkeywords}

\section{Introduction}

Conventionally, sensing and communication systems were designed separately for domain-specific tasks \cite{Qaisar26ISAC}. Due to the increasing importance of sensing in communication networks, next-generation wireless systems must support scenarios where both functionalities operate in a unified system. Motivated by this, integrated sensing and communication (ISAC) has attracted substantial attention from academia and industry over the past decade \cite{Han26IoT}. ISAC technologies are expected to enable a wide range of applications, including the Internet of Things (IoT), smart manufacturing and vehicular networks \cite{Qaisar26ISAC}.

To achieve adjustable sensing--communication tradeoffs or optimal joint performance under limited resources, advanced methods for resource allocation 
are required. In \cite{He23ISAC}, joint downlink beamforming and uplink power allocation problems are formulated to optimize the sum-rate or power consumption of a full-duplex ISAC system. 
In \cite{Tagliaferri24isac}, the constraints on communication and sensing signal-plus-distortion-and-noise ratios are satisfied via time–frequency power allocation. 
In \cite{Li2026V2X}, the objective is to jointly minimize communication multi-user interference, sensing waveform similarity and the Cram\'{e}r--Rao bound via space-time waveform design. 
These problems are usually nonconvex and are therefore decomposed into subproblems \cite{He23ISAC, Li2026V2X} and/or cast into well-studied optimization frameworks \cite{Tagliaferri24isac} to reduce complexity.

Despite the progress in existing works, conventional optimization algorithms often incur high computational complexity, particularly for high-dimensional and strongly nonconvex ISAC resource allocation problems. To further reduce complexity, machine learning-based approaches have attracted research interest \cite{W2P3}. Among various architectures, graph neural networks (GNNs) are especially promising for wireless network resource allocation, due to their ability to capture network topologies and generalize to unseen situations \cite{chen2024gnn, W2P2}.
Examples include \cite{W4P2, wang2024heterogeneous}, which address sum-rate maximization problems subject to signal-to-interference-plus-noise ratio (SINR), signal-to-noise ratio and/or transmit power-budget constraints using heterogeneous GNNs. Different types of transmitters, receivers, targets and their connected links are represented using different node and/or edge types, enabling optimal service modes \cite{W4P2} or beamforming \cite{wang2024heterogeneous}.
However, existing GNN-based ISAC approaches primarily model propagation-related interactions, such as communication channels and inter-link interference, without representing the power-budget coupling among links sharing a transmitter separately from interference \cite{W4P2, lian2026}.

To address this limitation, we propose a power-budget-aware graph representation with two distinct edge types: inter-group edges model same-channel interference, while intra-group edges capture power-budget coupling between links sharing a budget. This design lets messages pass even between links that share a budget but do not interfere. In addition, whereas existing works typically consider small, fixed network configurations \cite{wang2024heterogeneous, lian2026}, we investigate the scalability and generalization of the proposed GNN, termed GNN-ISAC, across various numbers of providers and links, including a model trained in one configuration and tested on larger ones without retraining or architectural modification. 
We also propose a benchmark algorithm by modifying the widely used power allocation method, weighted mean-square error minimization (WMMSE)~\cite{shi2011iteratively}, to incorporate ISAC SINR constraints.

\section{System Model and Problem Formulation}\label{sec:system}

\subsection{System Model}\label{sec:sysmodel}

\begin{figure}[h]
\centering
\includegraphics[width=\linewidth]{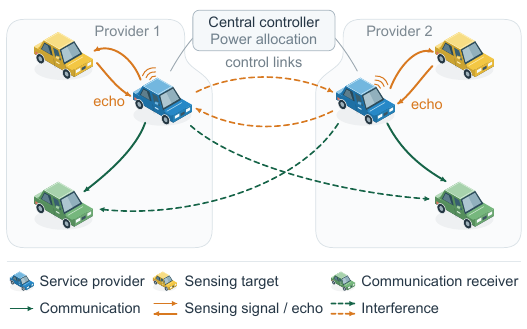}
\caption{Illustrative vehicular ISAC network with $B=2$.}
\label{fig:ISAC_scenario}
\end{figure}

As shown in Fig.~\ref{fig:ISAC_scenario}, we consider a vehicular ISAC network with a set $\mathcal{B}=\{1,\dots,B\}$ of service-provider vehicles. Provider $b\in\mathcal{B}$ serves a set $\mathcal{L}_{\mathrm{s}}^{b}$ of $M_{\mathrm{s}}^{b}$ sensing targets and a set $\mathcal{L}_{\mathrm{c}}^{b}$ of $M_{\mathrm{c}}^{b}$ communication receivers. We refer to each served target or receiver as a link, and write
$\mathcal{L}^{b}=\mathcal{L}_{\mathrm{s}}^{b}\cup\mathcal{L}_{\mathrm{c}}^{b}$
for the $M^{b}=M_{\mathrm{s}}^{b}+M_{\mathrm{c}}^{b}$ links of provider $b$. The set of all $K=\sum_{b\in\mathcal{B}}M^{b}$ links is $\mathcal{L}=\mathcal{L}_{\mathrm{s}}\cup\mathcal{L}_{\mathrm{c}}$, where $\mathcal{L}_{\mathrm{s}}=\bigcup_{b}\mathcal{L}_{\mathrm{s}}^{b}$ and $\mathcal{L}_{\mathrm{c}}=\bigcup_{b}\mathcal{L}_{\mathrm{c}}^{b}$. Each provider transmits data to its communication receivers and, simultaneously, transmits sensing signals to its sensing targets and collects the reflected signals. Hence, for a sensing link $k\in\mathcal{L}_{\mathrm{s}}^{b}$, provider $b$ is both the transmitter and the receiver.

The system has $M=\max_b M_{\mathrm s}^b+\max_b M_{\mathrm c}^b$ orthogonal channels indexed by $\mathcal M=\{1,\dots,M\}$. Each link $k$ is served by provider $b(k)$ on channel $m(k)\in\{1,\dots,M\}$. Within a provider, every link is assigned a different channel, and the assignment is the same for all providers, so each channel carries a single service type network-wide. Consequently, neither links of the same provider nor sensing and communication links interfere with each other. Since all providers reuse the $M$ channels, link $k$ is interfered by the links of other providers on the same channel,
\begin{equation}
\mathcal{I}_{k}=\{\,j\in\mathcal{L} : b(j)\neq b(k),\ m(j)=m(k)\,\}.
\end{equation}
Links of the same provider share a power budget $P_{\max}$ but do not interfere, whereas links on the same channel interfere but do not share a budget.

For a link $k \in \mathcal{L}$, the signal received by the intended receiver in the $m(k)$-th channel is given by 
\begin{equation}
y_{k}^{m(k)}=h_{k,b(k)}^{m(k)}\,s_{b(k)}^{m(k)}
+\sum_{j\in\mathcal{I}_{k}}h_{k,b(j)}^{m(k)}\,s_{b(j)}^{m(k)}+n_{k}^{m(k)},
\end{equation}
where $s_{b}^{m}$ is the signal transmitted by provider $b$ on channel $m$ with power $p_{b}^{m}=\mathbb{E}\{|s_{b}^{m}|^{2}\}$, $h_{k,b}^{m}\in\mathbb{C}$ is the channel state information (CSI) from provider $b$ to the receiver of link $k$ on channel $m$, and $n_{k}^{m}\sim\mathcal{CN}(0,\sigma^{2})$ is additive white Gaussian noise (AWGN). For a communication link $k \in \mathcal{L}_\mathrm{c}$, $h_{k,b(k)}^{m(k)}$ is the channel from provider $b(k)$ to its intended communication receiver; for a sensing link $k \in \mathcal{L}_\mathrm{s}$, it is the round-trip channel between the provider $b(k)$ and the target, and the intended receiver is provider $b(k)$ itself. Since each provider uses each channel for at most one link, $s_{b}^{m}$ is uniquely associated with a single link.
We assume quasi-static channels within each layout, with gains known to a central controller that allocates the powers, and neglect self-interference at the provider and interference scattered by the targets. Channel models are specified in Section~\ref{sec:sim-setup}.

The transmit powers form the power allocation matrix
$\mathbf{P}=[\mathbf{p}_1,\dots,\mathbf{p}_B]^{\top}\in\mathbb{R}^{B\times M}$,
where $\mathbf{p}_b=[p_{b}^{1},\dots,p_{b}^{M}]^{\top}$, and the SINR of link $k$ is
\begin{equation}
\gamma_{k}(\mathbf{P})=\frac{\bigl|h_{k,b(k)}^{m(k)}\bigr|^{2}p_{b(k)}^{m(k)}}
{\sum_{j\in\mathcal{I}_{k}}\bigl|h_{k,b(j)}^{m(k)}\bigr|^{2}p_{b(j)}^{m(k)}
+\sigma^{2}}.  \label{eq:sinr}
\end{equation}
The data rate of a communication link $k \in \mathcal{L}_{\mathrm{c}}$ is
\begin{equation}
R_{k}(\mathbf{P})=\log_2 (1+\gamma_{k}(\mathbf{P}))\quad\text{(bit/s/Hz)},
\end{equation}
while the SINR $\gamma_k(\mathbf P)$ of a sensing link $k\in\mathcal L_{\mathrm s}$ measures its sensing quality.





\subsection{Power Allocation Problem}




The objective of the ISAC power allocation problem is to find the power allocation matrix $\mathbf{P}^{\ast}$ that maximizes the communication sum-rate subject to the power-budget and sensing constraints, i.e.,
\begin{subequations}
\begin{align}
\mathbf{P}^{\ast} = \underset{\mathbf{P}}{\operatorname{argmax}}\quad
&\sum_{k\in \mathcal{L}_{\mathrm{c}}}   R_{k}(\mathbf{P})   \label{eq:prob_obj_sumrate}  \\ 
\operatorname{subject\ to}\quad
&\sum_{m \in \mathcal{M}}p_b^m\le P_{\max}, &&b \in \mathcal{B},  \label{eq:prob_const_powerbudget} \\
&\gamma_k( \mathbf{P})\ge\gamma_{\min}, &&k\in \mathcal{L}_{\mathrm{s}},   \label{eq:prob_const_sens} \\
&p_b^m\ge0, &&b\in\mathcal{B}, m \in \mathcal{M},   \label{eq:prob_const_non-neg}
\end{align}
\label{eq:ISAC_problem}%
\end{subequations}
where $\gamma_{\min}$ is the minimum SINR required by each sensing link before any radar processing gain. Constraint \eqref{eq:prob_const_powerbudget} ensures that for each provider, the total power allocated to its links does not exceed the power budget. Constraint \eqref{eq:prob_const_sens} guarantees the SINR of each sensing link. Constraint \eqref{eq:prob_const_non-neg} ensures that the transmit power allocated to each link is nonnegative. Due to the interference between the communication links, the problem is nonconvex and hard to solve. In the following sections, we first propose a modified WMMSE algorithm \cite{shi2011iteratively}, followed by an efficient GNN-based algorithm to solve the problem in \eqref{eq:ISAC_problem}.


\section{WMMSE with Sensing-Power Reservation }\label{ISAC-adapted-wmmse}  

The standard WMMSE method is designed to solve sum-rate optimization problems with power constraints. To tackle the additional per-sensing-link SINR constraints in \eqref{eq:prob_const_sens}, in this section we adapt WMMSE to this problem by introducing sensing-power reservation. More specifically, since sensing and communication do not interfere with each other as mentioned in Section~\ref{sec:sysmodel}, the modified WMMSE first solves \eqref{eq:prob_const_sens} to find minimum sensing link powers. The minimum sensing power is then subtracted from the total power budget, and the standard WMMSE is applied to the remaining communication sum-rate problem.

Substituting \eqref{eq:sinr} into \eqref{eq:prob_const_sens}, \eqref{eq:prob_const_sens} can be re-expressed as
\begin{equation}
\mathbf p_{\mathrm s}\ge\mathbf F\mathbf p_{\mathrm s}+\boldsymbol{\eta}, \label{eq:sens_const_vec}
\end{equation}
where $\mathbf p_{\mathrm s} \in \mathbb R_{+}^{| \mathcal{L}_{\mathrm{s}} | \times 1}$
is the vector of powers allocated to all sensing links,
$\mathbf{F} = (F_{kj}) \in \mathbb{R}_{+}^{|\mathcal{L}_{\mathrm{s}}| \times |\mathcal{L}_{\mathrm{s}}|}$
with $F_{kj} = \gamma_{\min}\bigl|h_{k,b(j)}^{m(k)}\bigr|^{2}/\bigl|h_{k,b(k)}^{m(k)}\bigr|^{2}$
for $j\in\mathcal I_k$ and $F_{kj}=0$ otherwise, with $k,j\in\mathcal L_{\mathrm s}$,
$\boldsymbol{\eta} = (\eta_k) \in \mathbb{R}_{+}^{|\mathcal{L}_{\mathrm{s}} | \times 1}$
with $\eta_k = \gamma_{\min} \sigma^2 / \bigl|h_{k,b(k)}^{m(k)}\bigr|^{2}$, and the vector inequality denotes the element-wise inequality.
When $\rho(\mathbf F)<1$, where $\rho(\mathbf F)$ denotes the largest eigenvalue of $\mathbf{F}$, the element-wise minimum sensing-power solution to \eqref{eq:sens_const_vec} is \cite{Yates95}
\begin{equation}
\mathbf p_{\mathrm s, \min}=(\mathbf{I}_{|\mathcal{L}_{\mathrm{s}}|} -\mathbf F)^{-1}\boldsymbol\eta, \label{eq:p_s_min}
\end{equation}
where $\mathbf{I}_{|\mathcal{L}_{\mathrm{s}}|}$ is a $|\mathcal{L}_{\mathrm{s}}|$-by-$|\mathcal{L}_{\mathrm{s}}|$ identity matrix.
Subtracting the minimum sensing powers in \eqref{eq:p_s_min} from $P_{\max}$, the remaining budgets for communication links can be expressed as
\begin{equation}
P_{\mathrm{c},b}=P_{\max}-\sum_{k \in\mathcal L_{\mathrm s}^b}(p_b^{m(k)})_{\min}, \quad b \in \mathcal{B},
\end{equation}
where $(p_b^{m(k)})_{\min}$ for $k \in\mathcal L_{\mathrm s}^b$ is the power of sensing link $k$ obtained from \eqref{eq:p_s_min}. The remaining communication power allocation problem is to maximize the sum-rate in \eqref{eq:prob_obj_sumrate} over nonnegative communication powers, subject to $\sum_{k\in\mathcal L_{\mathrm c}^{b}}p_b^{m(k)}\le P_{\mathrm{c},b}$ for all $b\in\mathcal B$. We apply the standard WMMSE approach \cite{shi2011iteratively} to this nonconvex subproblem, using equal communication-power shares within each provider as the initial point. The iterations stop when both rate-change and power-stationarity tolerances are satisfied, or when a predefined iteration limit is reached.

The original problem is feasible if and only if $\rho(\mathbf F)<1$ and $P_{\mathrm{c},b}\ge0, \forall b\in\mathcal B$. If no feasible allocation is available for a layout, equal power sharing is used among the links of every provider. This fallback satisfies the power-budget constraints but may violate sensing constraints.

\section{Graph Neural Networks for ISAC}\label{sec:gnn}

In this section, we present the graph representation for our problem and then propose a GNN-based algorithm that allocates transmit power for both sensing and communication links to maximize the communication sum-rate.

\subsection{Graph Representation}

Although an iterative WMMSE algorithm was introduced in the previous section to solve the problem, its computational complexity remains too high for practical implementation. To address this challenge, we develop a GNN-based approach to improve efficiency. A graph $\mathcal{G} = (\mathcal{V},\mathcal{E})$ can be characterized by the set of vertices $\mathcal{V}$ and edges $\mathcal{E}$ of a graph. For any given vertices $i,j\in\mathcal{V}$, $e_{i,j}$ defines the directed edge from $j$ to $i$.

We represent the network as a directed graph with sensing and communication links as vertices (nodes), and interference and power-budget relations as edges. We associate one vertex with each link, so that $|\mathcal{V}|=K$.
Since each provider assigns its links to distinct channels and the assignment is identical across providers, a link is uniquely identified by its provider and its channel. We therefore denote the vertex of the link served by provider $b$ on channel $m$ by $v_{b,m}^{\mathrm{c}}$ if channel $m$ carries communication, and by $v_{b,m}^{\mathrm{s}}$ if it carries sensing, so that $\mathcal{V}=\{\,v_{b(k),m(k)} : k\in\mathcal{L}\,\}$. We show the superscript only where the service type matters. Where a compact index is more convenient, we write $v_{k}\equiv v_{b(k),m(k)}$. Vertex $v_k$ represents the direct link from provider $b(k)$ to its target or receiver on channel $m(k)$. Its initial feature vector is
$\mathbf{x}_{k}^{(0)}=\bigl[\,\widetilde{d}_{kk},\ \widetilde{h}_{kk},\ t_{k}\,\bigr]$,
where $\widetilde{d}_{kk}$ is the normalized distance between provider $b(k)$ and the target or receiver of link $k$, $\widetilde{h}_{kk}$ is the normalized direct channel gain $\bigl|h_{k,b(k)}^{m(k)}\bigr|^{2}$, and $t_{k}$ indicates the service type, with $t_{k}=1$ if $k\in\mathcal{L}_{\mathrm{c}}$ and $t_{k}=0$ if $k\in\mathcal{L}_{\mathrm{s}}$. The geometry and channel features are standardized logarithmic transforms of inverse distance and channel power gain, respectively. For a sensing vertex, $\widetilde{h}_{kk}$ is derived from the round-trip gain.
The edge set is the union of two disjoint subsets, $\mathcal{E}=\mathcal{E}^{\mathrm{I}}\cup\mathcal{E}^{\mathrm{P}}$.

\emph{Interference edges:} An interference edge $e_{b,b'}^{\mathrm{I}}(m)$ is directed from $v_{b',m}$ to $v_{b,m}$, where the subscripts index the providers and the argument indexes the shared channel:
\begin{equation}
\mathcal{E}^{\mathrm{I}}=\{\,e_{b,b'}^{\mathrm{I}}(m) :
v_{b,m},v_{b',m}\in\mathcal{V},\ b\neq b'\,\}.
\end{equation}
Since each channel carries a single service type, interference edges only connect vertices of the same type. The edges $e_{b,b'}^{\mathrm{I}}(m)$ and $e_{b',b}^{\mathrm{I}}(m)$ are distinct, representing opposite directions whose channel-gain features may differ.

\emph{Power-budget edges:} A power-budget edge $e_{m,m'}^{\mathrm{P}}(b)$ is directed from $v_{b,m'}$ to $v_{b,m}$, where the subscripts index the channels and the argument indexes the shared provider. Such edges connect every pair of distinct links served by the same provider,
\begin{equation}
\mathcal{E}^{\mathrm{P}}=\{\,e_{m,m'}^{\mathrm{P}}(b) :
v_{b,m},v_{b,m'}\in\mathcal{V},\ m\neq m'\,\},
\end{equation}
so that the $M^{b}$ vertices of provider $b$ form a complete subgraph. Unlike interference edges, these edges do not correspond to a propagation path. Instead, they encode the constraint $\sum_{m\in\mathcal{M}}p_{b}^{m}\le P_{\max}$, which couples the powers of these links through a common budget.

The feature of edge $e_{k,j}$, where $e_{k,j}$ denotes the edge of either type from $v_{j}$ to $v_{k}$, is $\mathbf{z}_{k,j}=\bigl[\,\widetilde{d}_{kj},\ \widetilde{h}_{kj},\ E_{k,j}\,\bigr]$,
where $E_{k,j}=0$ for $e_{k,j}\in\mathcal{E}^{\mathrm{I}}$ and $E_{k,j}=1$ for $e_{k,j}\in\mathcal{E}^{\mathrm{P}}$. For an interference edge, $\widetilde{d}_{kj}$ is the normalized distance from provider $b(j)$ to the receiver of link $k$, and $\widetilde{h}_{kj}$ is the normalized interference gain $\bigl|h_{k,b(j)}^{m(k)}\bigr|^{2}$. For a power-budget edge, we set $\widetilde{d}_{kj}=\widetilde{h}_{kj}=0$. Accordingly, vertex $v_{b,m}$ has two neighbor sets: the interference neighbors
$\mathcal{N}^{\mathrm{I}}(v_{b,m})=\{v_{b',m}\in\mathcal{V} : b'\neq b\}$,
which corresponds to $\mathcal{I}_{k}$ for $v_{k}=v_{b,m}$, and the power-budget neighbors
$\mathcal{N}^{\mathrm{P}}(v_{b,m})=\{v_{b,m'}\in\mathcal{V} : m'\neq m\}$.

\begin{figure}[htbp]
  \centering
  \includegraphics[width=\linewidth]{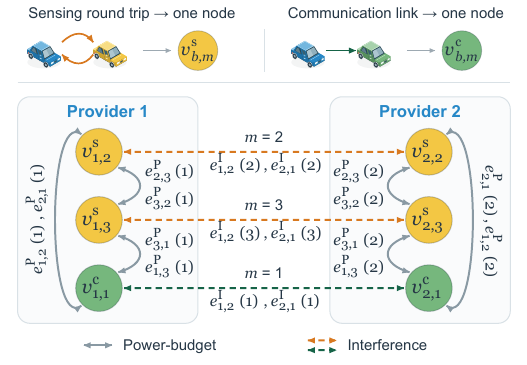}
  \caption{Link-node graph for $B=2$, $M_{\mathrm s}^b=2$ and $M_{\mathrm c}^b=1$.}
  \label{fig:network_to_graph}
  \label{fig:graph}
\end{figure}

For example, if each provider serves one communication receiver on channel~1 and two sensing targets on channels~2 and~3, provider $b$ contributes the vertices $v_{b,1}^{\mathrm{c}}$, $v_{b,2}^{\mathrm{s}}$ and $v_{b,3}^{\mathrm{s}}$, connected by the power-budget edges $e_{1,2}^{\mathrm{P}}(b)$, $e_{1,3}^{\mathrm{P}}(b)$ and $e_{2,3}^{\mathrm{P}}(b)$ together with their reverses, while the sensing vertices $v_{1,2}^{\mathrm{s}}$ and $v_{2,2}^{\mathrm{s}}$ of providers~1 and~2 are connected by $e_{1,2}^{\mathrm{I}}(2)$ and $e_{2,1}^{\mathrm{I}}(2)$, as illustrated in Fig.~\ref{fig:graph}.

\subsection{Message Passing and Power Allocation}\label{sec:gnn-message-passing}

The GNN allocates power by exchanging messages between the vertices in $\mathcal{V}$. Vertex $k$ maintains a scalar state $u_k^{(\ell)}$ after round $\ell$, with $u_k^{(0)}=0$ for each input graph. The final state serves as its power-allocation score. Its representation $\mathbf v_k^{(\ell)}=[\mathbf x_k^{(0)},u_k^{(\ell)}]$ concatenates the fixed input features with this state. Messages arrive from
$\mathcal N(k)=\mathcal N^{\mathrm I}(k)\cup\mathcal N^{\mathrm P}(k)$,
covering both interfering links and links sharing the provider's budget.

\begin{figure}[ht]
  \centering
  \includegraphics[width=\linewidth]{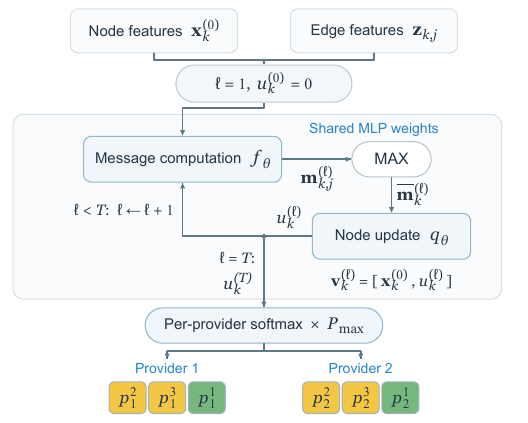}
  \caption{The structure of the proposed GNN-ISAC algorithm. }
  \label{fig:unrolled_layers}
\end{figure}

At each of the $T$ rounds, indexed by $\ell=1,\dots,T$, a message multilayer perceptron (MLP), $f_{\theta}$, combines the sender's representation with the edge features $\mathbf z_{k,j}$. Vertex $k$ aggregates the incoming messages and updates its state through a second MLP, $q_{\theta}$:
\begin{subequations}
\begin{align}
\mathbf m_{k,j}^{(\ell)}
&=f_{\theta}\!\left([\mathbf v_j^{(\ell-1)},\mathbf z_{k,j}]\right),
\qquad j\in\mathcal N(k),\\
\overline{\mathbf m}_k^{(\ell)}
&=\underset{j\in\mathcal N(k)}{\operatorname{MAX}}\,
\mathbf m_{k,j}^{(\ell)},\\
u_k^{(\ell)}
&=q_{\theta}\!\left([\mathbf v_k^{(\ell-1)},\overline{\mathbf m}_k^{(\ell)}]\right).
\end{align}
\end{subequations}
Here, $\mathbf m_{k,j}^{(\ell)}$ is the message from $j$ to $k$, and $\operatorname{MAX}$ takes the maximum of each component across all incoming messages. The resulting vector $\overline{\mathbf m}_k^{(\ell)}$ is combined with vertex $k$'s previous representation to update $u_k^{(\ell)}$. The parameter vector $\theta$ contains the trainable weights and biases of both MLPs; each MLP shares its own parameters across vertices, edge types and all $T$ rounds. The message MLP uses rectified linear unit (ReLU) activations; the update MLP has a ReLU hidden layer and a linear output. Only $u_k$ evolves across the rounds. After $T$ rounds, a per-provider softmax converts the final states into transmit powers:
\begin{equation}
p_{b(k)}^{m(k)}
=P_{\max}\frac{\exp(u_k^{(T)})}
{\sum_{j\in\mathcal L^{b(k)}}\exp(u_j^{(T)})},
\qquad k\in\mathcal L. \label{eq:softmax}
\end{equation}
The denominator includes both sensing and communication links. The resulting powers are nonnegative and sum to $P_{\max}$ for each provider, so the GNN uses the full budget although the problem permits unused power. Fig.~\ref{fig:unrolled_layers} illustrates this allocation process. The message and update MLPs each reuse their parameters over the $T$ rounds. Each vertex's scalar state starts at zero and is updated using its current state and the element-wise maximum of incoming messages from both edge types. The three input node features and the edge features remain fixed. Per-provider softmax converts final sensing and communication logits jointly into power fractions; multiplying these fractions by $P_{\max}$ gives the link powers.
\subsection{Unsupervised Training}

The GNN is trained without target power allocations. For a minibatch of $N$ layouts of one configuration, let $\mathbf P_n(\theta)$ be the predicted power allocation matrix for layout $n$ and $\gamma_{n,k}=\gamma_k(\mathbf P_n(\theta))$. Using the physical channel gains and Section~\ref{sec:system}'s rate and SINR expressions, we minimize
\begin{equation}
\begin{aligned}
\mathcal J(\theta)
={}&-\frac{1}{N}\sum_{n=1}^{N}\sum_{k\in\mathcal L_{\mathrm c}}
R_k\!\left(\mathbf P_n(\theta)\right)\\
&+\frac{\beta}{N}\sum_{n=1}^{N}\sum_{k\in\mathcal L_{\mathrm s}}
\left[\Gamma_{\min}-10\log_{10}(\gamma_{n,k}+\epsilon)\right]_+^2,
\end{aligned} \label{eq:loss}
\end{equation}
where $\Gamma_{\min}=10\log_{10}\gamma_{\min}$ is the sensing threshold in dB, $[a]_+=\max(a,0)$, $\beta>0$ weights the sensing penalty, and $0<\epsilon<\gamma_{\min}$ avoids evaluating the logarithm at zero. The loss combines the negative communication sum-rate with a squared sensing shortfall in dB, summing over links and averaging over layouts. 
A finite penalty weight $\beta$ does not guarantee that every sensing constraint is satisfied.
\begin{figure*}[t]
\centering
\includegraphics[width=1\textwidth]{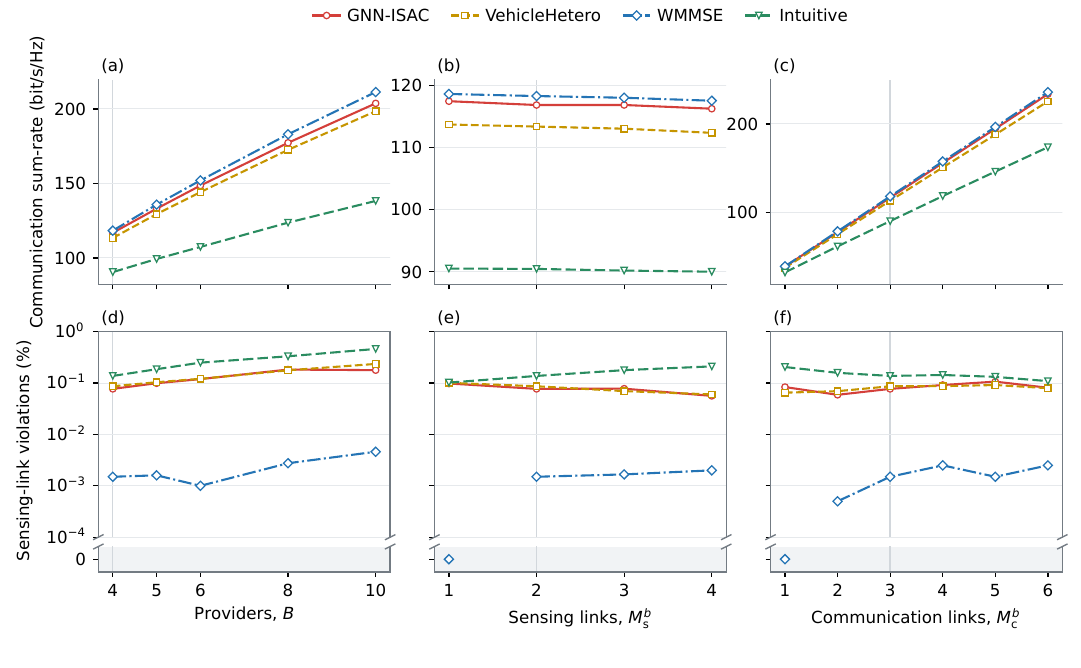}
\caption{Communication sum-rate (top) and sensing-link violations (bottom) versus $B$, $M_{\mathrm s}^b$ and $M_{\mathrm c}^b$ (left to right).} 
\label{fig:eval-quality}
\end{figure*}
During training, Adam updates $\theta$ after each minibatch using gradients obtained by backpropagation. Feature normalization uses statistics fitted only to training layouts. At inference, the model computes fresh node states and powers for each new graph through the operations in Section~\ref{sec:gnn-message-passing}. 

\section{Performance Evaluation}\label{C5:sec:performance_srm}

\subsection{Simulation Setup}\label{sec:sim-setup}

We consider service providers randomly placed in a $225\times225~\mathrm{m}^2$ region with a minimum separation of $50~\mathrm{m}$. Each sensing target or communication receiver is placed at a uniformly sampled distance of $2\text{–}5~\mathrm{m}$ and angle around its provider, subject to the region boundary and a minimum separation of $2~\mathrm{m}$ within the provider's served group. The carrier frequency is $2.4~\mathrm{GHz}$, the bandwidth is $5~\mathrm{MHz}$ per channel, and the provider power budget is $P_{\max}=10~\mathrm{W}$. The noise power is $\sigma^2=6.29\times10^{-14}~\mathrm{W}$ per channel, and the aggregate antenna gain for desired links is $2.5~\mathrm{dB}$. We adopt the communication channel model used in \cite{W2P3} and the sensing channel model from \cite{W4P2}. The sensing SINR threshold is $\Gamma_{\min}=-31.77~\mathrm{dB}$.






The proposed GNN-ISAC uses $T=4$, with layer dimensions $\{7, 32, 32\}$ for $f_\theta$ and $\{36,32,1\}$ for $q_\theta$. GNN-ISAC and VehicleHetero both use the loss in \eqref{eq:loss} with $\beta=0.05$ and $\epsilon=10^{-6}$. Training uses Adam for $30$ epochs with batch size $64$ and initial learning rate $0.002$, halved every $10$ epochs.

We compare with three baselines listed below:

\begin{itemize}
\item \textbf{WMMSE:} the modified WMMSE in Section~\ref{ISAC-adapted-wmmse} with sensing-power reservation and an iteration limit of $1,000$.
\item \textbf{VehicleHetero:} a heterogeneous GNN adapted from \cite{W4P2} using a vehicle-based graph representation and relation-specific mean aggregation.
\item \textbf{Intuitive:} an intuitive method that reserves a fixed share of power for sensing links, and divides equally among both sensing and communication partitions separately.
\end{itemize}

For the main comparisons, each learned method is trained separately in each configuration using $10,000$ training, $1,000$ validation and $5,000$ test layouts per seed, over five independent seeds.

\subsection{Communication Performance}

The nominal configuration is $(B,M_{\mathrm s}^b,M_{\mathrm c}^b)=(4,2,3)$ for every provider. We vary $B\in\{4,5,6,8,10\}$ in Figs.~\ref{fig:eval-quality}(a), (d), $M_{\mathrm s}^b\in\{1,2,3,4\}$ in Figs.~\ref{fig:eval-quality}(b), (e) and $M_{\mathrm c}^b\in\{1,2,3,4,5,6\}$ in Figs.~\ref{fig:eval-quality}(c), (f), respectively, holding the other parameters at their nominal values.

Figs.~\ref{fig:eval-quality}(a)--(c) compare the communication sum-rate under different numbers of providers, sensing links and communication links. GNN-ISAC achieves $96.4\text{–}99.4\%$ of the WMMSE rate across the displayed configurations and consistently outperforms VehicleHetero and Intuitive. Its graph provides direct-channel node features, interference-channel edge features and shared-budget relations, so the allocation can respond jointly to channel gains, interference and power competition. Intuitive instead draws channel-independent power shares.

The sweep trends reflect changes in resources and interference. Increasing either $B$ or $M_{\mathrm c}^b$ raises the sum-rate. Each added provider brings communication links but also interferers, whereas increasing $M_{\mathrm c}^b$ adds communication links on new orthogonal channels, so the sum-rate grows faster per added link. Increasing $M_{\mathrm s}^b$ affects communication signal and interference powers in the same way by sharing the power budgets, thus having limited impact on the sum-rate.



\subsection{Sensing Reliability}

Figs.~\ref{fig:eval-quality}(d)--(f) show the sensing-link violation rate, i.e., the percentage of sensing links with $\gamma_k(\mathbf P)<\gamma_{\min}$, which is $0.056\text{–}0.183\%$ for GNN-ISAC. The two learned methods have similar violation levels, while WMMSE provides stronger sensing-constraint satisfaction. This difference reflects how the methods handle sensing requirements. WMMSE explicitly reserves sensing power and checks its solver allocations against the constraints. GNN-ISAC instead learns to balance communication rate and sensing shortfall through a finite penalty and does not enforce the sensing thresholds.

\subsection{Generalization}

We first train GNN-ISAC on a small network, i.e., $(B,M_{\mathrm s}^b,M_{\mathrm c}^b) = (4,2,3)$. Then, we test the trained network on larger configurations without retraining or refitting. The generalization performance is shown in Table~\ref{tab:eval-generalisation}. Performance is normalized by the sum-rate of the configuration-specific trained GNN-ISAC at the same test layouts. Notably, it retains $95.5\%$ of the configuration-trained sum-rate even when the number of links is 3.5 times larger, i.e., $(B,M_{\mathrm s}^b,M_{\mathrm c}^b) = (10,3,4)$, highlighting its adaptability to different layout configurations.

\begin{table}[t]
\centering
\caption{Communication-rate retention of GNN-ISAC without retraining.}
\label{tab:eval-generalisation}
\begingroup
\footnotesize
\setlength{\tabcolsep}{1.5pt}
\renewcommand{\arraystretch}{1.18}
\begin{tabular*}{\linewidth}{@{\extracolsep{\fill}}ccccc@{}}
\hline
\begin{tikzpicture}[baseline=(current bounding box.center),x=1mm,y=1mm]
  \path[use as bounding box] (0,0) rectangle (26,11);
  \draw[line width=0.3pt] (0,11) -- (26,0);
  \node[anchor=center,inner sep=0pt] at ({26/3},{11/3}) {$B$};
  \node[anchor=north east,inner sep=0pt] at (25.3,10.3)
    {$(M_{\mathrm s}^{b},M_{\mathrm c}^{b})$};
\end{tikzpicture}
& $(2,3)$ & $(3,3)$ & $(2,4)$ & $(3,4)$ \\
\hline
$5$  & $99.0\%$ & $98.6\%$ & $99.0\%$ & $99.1\%$ \\
$6$  & $97.6\%$ & $97.9\%$ & $97.9\%$ & $97.9\%$ \\
$7$  & $97.0\%$ & $96.9\%$ & $97.1\%$ & $97.0\%$ \\
$8$  & $96.4\%$ & $96.2\%$ & $96.6\%$ & $96.5\%$ \\
$10$ & $95.5\%$ & $95.4\%$ & $95.7\%$ & $95.5\%$ \\
\hline
\end{tabular*}
\endgroup
\end{table}

\subsection{Computational Cost}

\begin{figure}[t]
\centering
\includegraphics[width=\linewidth]{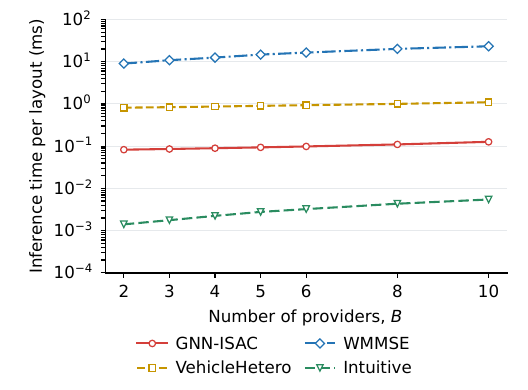}
\caption{Computation time versus provider count.}
\label{fig:eval-runtime}
\end{figure}

Fig.~\ref{fig:eval-runtime} compares the computation time for $B\in\{2,3,4,5,6,8,10\}$ with $(M_{\mathrm s}^b,M_{\mathrm c}^b)=(2,3)$. At $B=10$, the computation times are $0.127~\mathrm{ms}$ for GNN-ISAC and $23.22~\mathrm{ms}$ for WMMSE. GNN-ISAC learns the graph-to-power mapping offline. At inference, four message-passing rounds followed by the per-provider softmax replace WMMSE's repeated communication-power updates for each new layout. The number of rounds stays fixed as the network grows, although each round must process more vertices and edges. Intuitive is cheaper, but its channel-independent rule delivers much lower rates.

For repeated allocation decisions, the proposed \mbox{GNN-ISAC} combines near-WMMSE communication performance and low empirical sensing violations with substantially lower online computation.

\section{Conclusion}

In this paper, we studied power allocation in vehicular ISAC networks, where each service-provider vehicle shares its power budget between sensing and communication links. We proposed a GNN-based algorithm whose graph uses two edge types to represent same-channel interference and shared power budgets, and a WMMSE benchmark modified with sensing-power reservation. Simulation results show that the proposed GNN-ISAC attains near-WMMSE communication sum-rates with low sensing-link violation rates and much less computation. Moreover, GNN-ISAC trained on a small network retains most of its sum-rate on larger networks without retraining, highlighting its potential for repeated power allocation across new network layouts.

\vspace{12pt}

\end{document}